\documentclass[letterpaper, 10 pt, conference]{style/ieeeconf}  % Comment this line out if you need a4paper

\IEEEoverridecommandlockouts % This command is only needed if 
\usepackage{amsmath}
\usepackage{amssymb}
\usepackage{graphicx}
\usepackage{booktabs}
\usepackage{bbm}
\usepackage{multirow}
\usepackage{graphicx}
\usepackage{amsmath}
\usepackage{amssymb}
\usepackage{tabularx}
\usepackage{array}
\usepackage{cuted}
\usepackage{capt-of}
\usepackage{url}
\usepackage{wrapfig}
\usepackage{hyperref}

\usepackage{xcolor}
\usepackage{listings}
\usepackage[most]{tcolorbox}

\definecolor{capgreen}{RGB}{115,160,35}
\definecolor{capblue}{RGB}{224,235,250}
\definecolor{capborder}{RGB}{190,190,190}

\lstdefinestyle{capcode}{
language=Python,
basicstyle=\ttfamily\scriptsize,
keywordstyle=\bfseries,
commentstyle=\color{capgreen},
backgroundcolor=\color{capblue},
frame=none,
showstringspaces=false,
columns=fullflexible,
keepspaces=true,
breaklines=true,
xleftmargin=2pt,
xrightmargin=2pt,
aboveskip=1pt,
belowskip=1pt
}

\newtcolorbox{capbox}{
colback=white,
colframe=capborder,
boxrule=0.4pt,
arc=0pt,
left=3pt,
right=3pt,
top=3pt,
bottom=3pt,
boxsep=0pt
}

\newcommand{\conf}[1]{\textcolor{gray}{\scriptsize [#1]}}
\definecolor{muyicolor}{RGB}{200,50,50}
\definecolor{jingfancolor}{RGB}{150,40,200}

\definecolor{gainblue}{HTML}{2F6BDE}
\newcommand{\gain}[1]{\textcolor{gainblue}{\scriptsize #1}}

\title{\LARGE \bf
Map2Route: Benchmarking Compositional Language-Grounded Route Planning over Semantic Maps
}

\author{
Muyi Bao$^{1}$, Hang Xu$^{1}$, Jingfan Tang$^{1}$, Zihan Liu$^{1}$, 
Yuxin Cai$^{2}$, Chen Lv$^{2}$, Wenshan Wang$^{1}$, Ji Zhang$^{1}$%
\thanks{$^{1}$Muyi Bao, Hang Xu, Jingfan Tang, Zihan Liu, Wenshan Wang, and Ji Zhang are with Carnegie Mellon University, Pittsburgh, PA 15213, USA.
E-mail: \{muyib, wx2, jingfant, ...\}@andrew.cmu.edu.}%
\thanks{$^{2}$Yuxin Cai and Chen Lv are with Nanyang Technological University, Singapore 639798
(e-mail: caiy0039@e.ntu.edu.sg; lyuchen@ntu.edu.sg).}
}

\begin{document}

\IEEEaftertitletext{%
\vspace{-4pt}
\begin{minipage}{\textwidth}
\centering
\includegraphics[width=\linewidth]{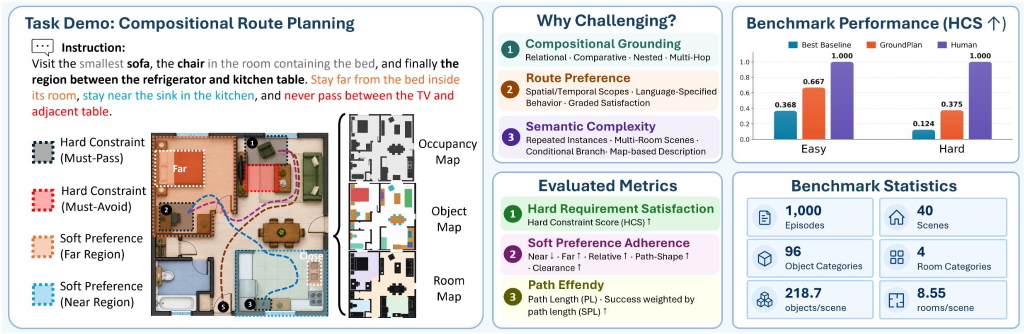}
\captionof{figure}{
Overview of Map2Route. Given a pre-built semantic map and a natural-language instruction, the task is to generate a complete route that resolves compositional references and follows ordered, scoped route requirements and preferences. 
Map2Route contains 1,000 episodes across 40 multi-room scenes and evaluates performance across hard constraints, soft-preference adherence, and path efficiency. 
Existing methods remain substantially below human performance, highlighting the challenge of compositional language-grounded route planning.
}
\label{fig:overview}
\end{minipage}
\vspace{2pt}
}

\maketitle
\thispagestyle{empty}
\pagestyle{empty}

%%%%%%%%%%%%%%%%%%%%%%%%%%%%%%%%%%%%%%%%%%%%%%%%%%%%%%%%%%%%%%%%%%%%%%%%%%%%%%%%

\begin{abstract}
% Planning from natural-language instructions over semantic maps involves more than identifying a navigation goal, as instructions may specify where to go through compositional references while also imposing scoped route preferences.
% Existing works typically capture only subsets of these capabilities, leaving their joint evaluation largely unexplored.
We introduce \textbf{Map2Route}, a human-curated benchmark for compositional language-grounded route planning over pre-built semantic maps.
Map2Route contains 1,000 episodes across 40 scenes, where instructions use relational, comparative, and nested descriptions to identify route-relevant objects and regions, while specifying ordered must-pass regions, must-avoid requirements, five categories of soft preferences, and spatial and route-stage scopes, which is partially tested by existing works.
Alongside Map2Route, we propose \textbf{Grounding2Route}, which combines executable code-as-grounding with verification-guided repair and scope-aware planning.
Across seven representative adapted baselines, Grounding2Route substantially outperforms existing methods in all metrics.
Despite these gains, a substantial gap to human demonstrations remains, highlighting the difficulty of Map2Route and the considerable headroom for future progress. 
Additional qualitative results and resources are available on \href{https://anonymous.4open.science/w/Map2Route-F05F/}{Project Page}.

% \jingfan{I feel like that the focus of the paper is on benchmark side. In abstract, you should strength the importance of this benchmark (why it is important? What ability of robot is tested? Did previous benchmark test it? Is robot not good enough for now?)}
\end{abstract}

%%%%%%%%%%%%%%%%%%%%%%%%%%%%%%%%%%%%%%%%%%%%%%%%%%%%%%%%%%%%%%%%%%%%%%%%%%%%%%%%

\section{Introduction}
From the long-standing vision of household companions to robots entering everyday workplaces, the aspiration of communicating with intelligent machines as naturally as with another person has long shaped the promise of robotics. In many practical settings, such as homes, offices, and factories, robots repeatedly operate in the same environments, where maps can be constructed in advance or maintained over long-term deployment and reused for localization and planning~\cite{VLMaps,iln,sysnav}. Once such spatial knowledge is available, users should not need to specify destinations through object IDs or absolute coordinates on a map. Instead, users should be able to use map-based natural-language instructions ~\cite{sayplan} (e.g. \textit{``go to the chair nearest to the bed in the bedroom with two windows''}) to describe the route an agent should follow.

This work considers \textbf{language-grounded route planning over pre-built semantic maps}. Given traversability, object, and room maps, and a natural-language instruction, the goal is to generate a complete route that faithfully realizes the instruction on the map.

Such language-grounded route planning raises three key challenges. 
\textbf{Challenge 1: compositional grounding.} 
The agent must determine which objects or regions in the map are referred to by natural language. Since multiple instances of the same category often exist, targets can be specified through relational, comparative, and nested references. For example, \textit{``the chair nearest to the bed in the bedroom with two windows that is farthest from the kitchen''} requires several reasoning steps to identify a single chair. 
\textbf{Challenge 2: soft-preference adherence.} 
Instructions describe not only \emph{where to go}, but also \emph{how to get there} through soft route preferences. For example, \textit{``stay far from the windows while crossing the bedroom''} requires the agent to understand the intended preference and generate a route that reflects it.
\textbf{Challenge 3: semantic complexity.} 
The agent must integrate grounded references and route requirements under substantial semantic complexity, including multi-room traversal, repeated object instances, multiple destinations (e.g., \textit{``go to the sofa, then the desk''}), ordering constraints (e.g., \textit{``before going to the desk, visit the chair''}), and conditional branching (e.g., \textit{``if the bedroom contains a lamp, go to the lamp; otherwise, go to the table''}).

Existing work has separately studied language-grounded goals and simple spatial relations~\cite{VLMaps} (\textit{Challenge 1}), behavioral route preferences~\cite{Behav,NORM-Nav} (\textit{Challenge 2}), and multi-stage task and temporal planning~\cite{sayplan,osgllm,lang2ltl,limp,ltlcodegen} (\textit{Challenge 3}), but rarely evaluates these capabilities jointly over complete routes, often with limited compositional complexity.

To evaluate jointly, we introduce \textbf{Map2Route}, a human-curated benchmark for compositional language-grounded and constraint-aware route planning over semantic maps. Map2Route evaluates three complementary dimensions: 1) hard-requirement satisfaction for compositional grounding and multi-stage route reasoning (\textit{Challenge 1}), since grounding errors ultimately lead to incorrect or violated route requirements; 2) soft-preference adherence for whether the route reflects language-specified preferences beyond simply reaching the destination (\textit{Challenge 2}); and 3) path efficiency for how efficiently the intended route is realized.

Alongside the benchmark, we propose a strong reference method, \textbf{Grounding2Route}, a structured language-to-route planning framework. 
Inspired by Code-as-Policy \cite{codeaspolicy}, Grounding2Route uses executable \emph{code-as-grounding} to produce a \emph{Route Semantic Intermediate Representation (RouteIR)}, which is deterministically compiled into a route through scope-aware planning.

On Map2Route, we evaluate seven adapted representative methods, Grounding2Route, shortest-feasible A*, and human demonstrations. The strongest baseline satisfies only 29.4\% of the \textit{hard-requirement satisfaction}, whereas Grounding2Route improves this to 57.9\%, while also outperforming all adapted baselines across all metrics. Despite these gains, the substantial gap to human demos shows that Map2Route remains highly challenging and far from saturated.

Our contributions are threefold:
\begin{itemize}
    \item Our primary contribution is \textbf{Map2Route}, a human-curated benchmark for compositional language-grounded route planning over semantic maps.

    \item To provide a strong reference method for this benchmark, we propose \textbf{Grounding2Route}, a two-stage structured language-to-route framework. 

    \item We provide a comprehensive empirical study showing that seven adapted representative methods struggle substantially on Map2Route, and conduct task-specific controlled ablations to identify the key components required for this task. We further analyze benchmark difficulty and validate the framework in real-world trials.
\end{itemize}

\section{Related Work}

\subsection{Language-Guided Planning over Pre-built Maps}

% Existing methods address different aspects of language-guided planning over structured spatial representations. VLMaps~\cite{VLMaps} grounds language-specified goals and spatial relations on semantic maps; SayPlan and OSG-LLM~\cite{sayplan,osgllm} reason over structured scene representations for task-level planning; temporal-logic methods~\cite{lang2ltl,lang2tlt2,limp,ltlcodegen} translate language into formal spatial and temporal specifications; and BehAV~\cite{Behav} and NORM-Nav~\cite{NORM-Nav} incorporate language-specified behaviors into geometric navigation objectives.

% However, each line of work covers only part of the full route-planning problem: VLMaps \cite{VLMaps} mainly addresses goal grounding and spatial relations; SayPlan \cite{sayplan} and OSG-LLM \cite{osgllm} emphasize task-level sequencing; temporal-logic methods focus on hard spatial, temporal, and ordering constraints; and BehAV \cite{Behav} and NORM-Nav \cite{NORM-Nav} primarily model behavioral preferences during navigation. Instead, Map2Route jointly evaluates compositional grounding, multi-stage route requirements, and soft preferences in a unified complete-route setting.

Existing methods address different aspects of language-guided planning over structured spatial representations. VLMaps~\cite{VLMaps} grounds language-specified goals and spatial relations onto semantic maps, enabling navigation to relationally defined locations. SayPlan and OSG-LLM~\cite{sayplan,osgllm} reason over structured scene representations to identify task-relevant entities and produce task-level or symbolic plans. Temporal-logic methods~\cite{lang2ltl,lang2tlt2,limp,ltlcodegen} translate natural-language instructions into formal spatial, temporal, or ordering specifications, while BehAV~\cite{Behav} and NORM-Nav~\cite{NORM-Nav} incorporate language-specified behavioral constraints into geometric navigation objectives.

However, these formulations emphasize different subsets of the full route-planning problem. VLMaps \cite{VLMaps} primarily focuses on grounding goals and spatial relations, while SayPlan \cite{sayplan} and OSG-LLM \cite{osgllm} emphasize task-level reasoning and action sequencing rather than detailed route preferences. Temporal-logic approaches \cite{lang2ltl,lang2tlt2,limp,ltlcodegen} are well suited to discrete hard requirements such as ordering or avoidance, but are less naturally suited to graded preferences over route geometry. Conversely, BehAV \cite{Behav} and NORM-Nav \cite{NORM-Nav} model behavioral preferences during navigation, but do not jointly address complex compositional grounding and multi-stage route semantics. In comparison, Map2Route jointly evaluates complex compositional grounding, multi-stage route requirements, and scoped soft preferences over complete routes.

\subsection{Online Language-Guided Navigation}

Online language-guided navigation requires an agent to interpret language while sequentially interacting with an environment from egocentric observations. Vision-and-Language Navigation (VLN)~\cite{r2r,rxr,goal2pixel} requires agents to follow natural-language route descriptions, while ObjectNav, REVERIE, and SOON~\cite{objnav,intentnav,reverie,soon} specify navigation goals through semantic categories or referring expressions. These tasks study online perception, grounding, and sequential decision making without access to a pre-built semantic map. Map2Route instead assumes that a semantic map is already available and studies how free-form language can be translated into a complete route over this map.

\section{Map2Route}
\label{sec:benchmark}
\begin{figure*}[h]
\centering
\includegraphics[width=\linewidth]{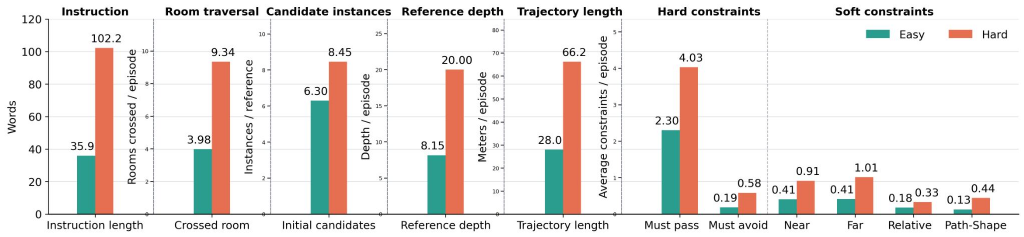}
\caption{Complexity statistics of the Easy and Hard subsets in Map2Route.
}
\label{fig:statistics}
\vspace{-6mm}
\end{figure*}

\subsection{Task Formulation}
\label{sec:task_formulation}

Each episode provides a semantic map $\mathcal{M}=(\mathcal{M}_{\mathrm{trav}},\mathcal{M}_{\mathrm{obj}},\mathcal{M}_{\mathrm{room}})$ comprising aligned traversability, object, and room layers, an initial robot position $p_0$, and a natural-language instruction $I$, but no explicit goal coordinates. The task is to generate a valid route
\begin{equation}
(\mathcal{M},p_0,I)
\;\longrightarrow\;
\tau=(p_0,p_1,\ldots,p_T)
\end{equation}
that faithfully realizes $I$ in $\mathcal{M}$. Map2Route therefore evaluates the complete generated route.

\subsubsection{Semantic Map Representation}
Each environment provides aligned traversability, object, and room maps at $0.05$\,m/grid resolution, generated offline from ProcTHOR~\cite{ProcTHOR} annotations and AI2-THOR~\cite{aithor} metadata. Traversability is obtained from reachable positions, object footprints from projected 3D bounding boxes, and room labels from rasterized room polygons. 
Walls are rasterized separately and non-square maps are padded as untraversable.

\subsubsection{Compositional Route Instructions}
Each instruction is decomposed into hard requirements and soft preferences. Hard requirements include ordered must-pass and must-avoid regions, while soft preferences have five types: Near, Far (be close to or far away from a reference object), Relative (be closer to one object than another), Path-Shape (follow a path of a certain shape), and Clearance (keep distance from obstacles) preferences, each defined with an active region. Language-specified constraints may further have spatial or temporal scopes; Clearance is applied globally by default. This decomposition is used only for annotation and evaluation, while methods receive the original instruction.

\subsection{Benchmark Construction}
\label{sec:benchmark_construction}

Map2Route contains 1,000 evaluation-only episodes across 40 ProcTHOR-10K~\cite{ProcTHOR} scenes, including 700 \textit{Easy} and 300 \textit{Hard} episodes, with 25 instructions per scene. We uniformly subsample 40 scenes from the 120 environments with the largest floor areas and use four additional disjoint scenes for a separate 50-episode development set. Two experts spent approximately 120 hours on annotation and cross-checking, followed by 20 hours of quality control for ambiguity, uniqueness and feasibility. Example episodes are available on the \href{https://anonymous.4open.science/w/Map2Route-F05F/}{Project Page}.

\textit{Annotation Principles.}
We follow three principles:
\textit{(1) Complexity.} \textit{Easy} episodes contain 2--4 constraints, while \textit{Hard} episodes contain 5--8 and are annotated afterward to increase compositional complexity.
\textit{(2) Diversity.} We encourage diverse objects, rooms, referring expressions, constraint types and combinations, and spatial or route-stage scopes.
The first two principles make \textit{Hard} episodes systematically more challenging than \textit{Easy} ones, with more constraints, more complex referring expressions and reference chains, and richer combinations of constraint types and scopes.
\textit{(3) Validity.} Each reference must uniquely identify a concrete object or region, and all specified routes must be feasible.

\textit{Benchmark Statistics.}
Fig.~\ref{fig:statistics} summarizes instruction length, room traversal, candidate instances before disambiguation, reference depth, trajectory length, and the numbers of hard and soft constraints. The benchmark further spans 96 object categories and 4 room types. Each scene contains an average of 218.7 object instances and 8.55 rooms. \textit{Hard} episodes show greater complexity than \textit{Easy} episodes across all reported dimensions.

% \subsection{Benchmark Metrics}
% \label{sec:benchmark_metrics}

% We evaluate along three complementary dimensions: 1) hard-requirement score, which measures whether a method can correctly resolve and satisfy language-specified destinations and ordered route requirements; 2) soft-preference adherence, which evaluates whether the generated route follows the specified preferences within their intended scopes; and (3) path efficiency, which measures the efficiency of the generated route.

\subsection{Benchmark Metrics}
\label{sec:benchmark_metrics}

We evaluate along three complementary dimensions: 1) hard-requirement score, which measures whether a method can resolve and satisfy language-specified destinations and ordered route requirements; 2) soft-preference adherence, which evaluates whether the route follows specified preferences within their intended scopes; and (3) path efficiency, which measures the efficiency of the generated route.

\subsubsection{Hard Constraint Score (HCS)}
A path can be considered containing an ordered list $\mathcal{G}$ of hard requirement regions $G_1,\ldots,G_{M+N}$, including M must-pass and N must-avoid region. These regions are specified either through language description (ex. area between the bed and the desk), or through a reference object. For a reference object $O_i$,
\begin{equation}
G_i =
\left\{
p\in R(O_i)
\;\middle|\;
\lVert p-c_{O_i}\rVert_2
\leq r_{O_i}+1\,\mathrm{m}
\right\},
\end{equation}
where $R(O_i)$ is its room and $c_{O_i},r_{O_i}$ denote its footprint center and radius. A region is reached when the route intersects it. HCS is defined as the ratio of the number of correctly reached must-pass regions under the order defined in $\mathcal{G}$, $K$, and the total number of must-pass regions, $M$:
\begin{equation}
\mathrm{HCS}=\frac{K}{M}.
\end{equation}
Note that a route-stage violation stops further credit, while a global violation yields $\mathrm{HCS}=0$.

\subsubsection{Soft Preference Metrics}
Let $\mathcal{W}_R(\tau)$ denote the ordered sequence of distinct route locations within active region $R$, where repeated visits to the same location are counted only once while preserving the order of first occurrence; $\mathcal{W}(\tau)$ is defined analogously over the full route. 
Let $d(p,O)$ and $d(p,\mathcal{U})$ denote the closest-point distance between a point $p$ to an object and the nearest non-traversable area, respectively. 
% Table~\ref{tab:soft_metrics} defines the five metrics. 
We select these five preference types because they cover a broad range of common route-shaping preferences, directly affect the geometry of the generated route, and remain sufficiently distinct from one another to avoid highly redundant evaluation.
Table~\ref{tab:soft_metrics} defines the corresponding metrics.
Since they capture different route preferences and have different scales and optimization directions, we report them separately rather than combining them into a single score. 
For each preference type, we first average valid constraint-level measurements within each episode and then macro-average across valid episodes.

\begin{table}[h]
\centering
\caption{Definition of soft constraint metrics.}
\label{tab:soft_metrics}

\resizebox{0.40\textwidth}{!}{%
\begin{tabular}{lcc}
\toprule
\textbf{Preference} & \textbf{Metric} & \textbf{Better} \\
\midrule

Near
& $\displaystyle
\frac{1}{|\mathcal{W}_R(\tau)|}
\sum_{p\in\mathcal{W}_R(\tau)}
d\!\left(p,O\right)$
& $\downarrow$ \\[5pt]

Far
& $\displaystyle
\frac{1}{|\mathcal{W}_R(\tau)|}
\sum_{p\in\mathcal{W}_R(\tau)}
d\!\left(p,O\right)$
& $\uparrow$ \\[5pt]

Relative
& $\displaystyle
\frac{D_B}{D_A + \epsilon}$
& $\uparrow$ \\[5pt]

Path-Shape
& $\displaystyle
\exp\!\left(
-\frac{
\operatorname{DTW}_{\delta}
\!\left(\mathcal{W}_R(\tau),\tau_{\mathrm{ref}}\right)}
{d_{\mathrm{succ}}|\tau_{\mathrm{ref}}|}
\right)$
& $\uparrow$ \\[5pt]

Clearance
& $\displaystyle
\frac{1}{|\tau|}
\sum_{p\in\tau}d(p,\mathcal{U})$
& $\uparrow$ \\

\bottomrule
\end{tabular}%
}

\end{table}

\begin{figure*}[h]
\centering
\includegraphics[width=1\linewidth]{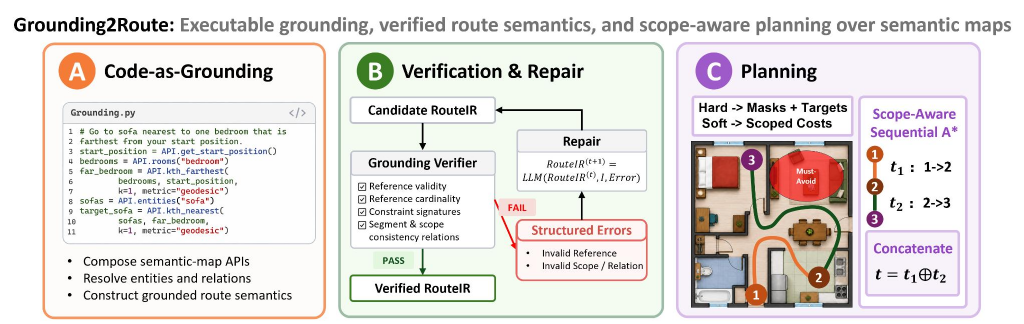}
\caption{
Overview of the Grounding2Route pipeline. Grounding2Route first uses executable code-as-grounding to resolve compositional route semantics over the semantic map, then verifies and repairs the resulting RouteIR, and finally compiles the verified specification into a complete route through deterministic scope-aware sequential planning.
}
\vspace{-6mm}
\label{fig:Grounding2Route}
\end{figure*}

For Relative, $A$ is the reference the route should stay closer to and $B$ is the one the route should stay farther from, with mean distances $D_A$ and $D_B$. Path-Shape uses tolerance-aware nDTW against a canonical trajectory $\tau_{\mathrm{ref}}$ ($\delta=0.5$\,m, $d_{\mathrm{succ}}=0.2$\,m). Independently drawn human trajectories score $0.99$ on both Easy and Hard, showing that the tolerance-aware metric assigns nearly identical scores to alternative valid paths and thus supports using a single canonical reference.
For Near, Far, Relative, and Path-Shape, failure to enter the target region or complete the relevant segment receives the metric-specific worst value and remains in the denominator; Clearance is always evaluated globally.

\subsubsection{Path Efficiency}
We report path length (PL) and Success weighted by Path Length (SPL),
\begin{equation}
\mathrm{SPL}
=
\displaystyle
\frac{1}{M}
\sum_{m=1}^{M}S_m\frac{L^*_m}{\max(L^*_m,L_m)},
\end{equation}
where $S_m=1$ if the agent ever reaches $G_{m}$ from $G_{m-1}$ and $0$ otherwise, $L_m$ is the agent's route length starting from $G_{m-1}$ until it reaches $G_{m}$ (if ever), and $L^*_m$ is the shortest feasible route produced by A* satisfying the same hard constraint connecting $G_{m-1}$ and $G_{m}$.

\section{Grounding2Route}
\label{sec:Grounding2Route}
We propose \textbf{Grounding2Route}, a structured language-to-route planning framework, as shown in Fig.~\ref{fig:Grounding2Route}. Grounding2Route consists of two stages: \emph{grounding} and \emph{route planning}. In the grounding stage, an LLM composes executable read-only semantic-map APIs through \emph{code-as-grounding} (Sec.~\ref{sec:grounding}), followed by verification-guided repair (Sec.~\ref{sec:verification}) to produce a grounded RouteIR (Sec.~\ref{sec:route_specification}). The verified RouteIR is then deterministically compiled into a complete route through scope-aware sequential planning (Sec.~\ref{sec:scope_planning}).

\subsection{Executable Code-as-Grounding}
\label{sec:grounding}

Given an instruction $I$ and semantic map $\mathcal{M}$, compositional references may require multi-step reasoning over relational, comparative, and nested descriptions. Inspired by Code-as-Policies~\cite{codeaspolicy}, we propose \textit{code-as-grounding (CaG)}, where an LLM composes predefined read-only semantic-map APIs and helper logic into a grounding program $P$, demoed in Fig.~\ref{fig:Grounding2Route}.A. Executing $P$ produces a provisional grounding

\begin{equation}
    RouteIR = P(\mathcal{M},p_0),
\end{equation}
where $RouteIR$ is a provisional route semantic intermediate representation, detailed in Sec.~\ref{sec:route_specification}. All map-dependent bindings in $G$ are computed during execution rather than directly predicted by the LLM.

The API provides primitives across six categories: scene/room queries, selection/set operations, ranking/spatial predicates, logic/comparison, region/waypoint construction, and route construction. 
These primitives can be compositionally combined to resolve nested references.
For Path-Shape, the generated code constructs ordered waypoints. Although Path-Shape is evaluated as a soft preference, Grounding2Route enforces the waypoint order during planning because additive spatial costs cannot reliably preserve topological ordering.

The generated program is sandboxed: it may compose exposed primitives through pure helper functions, but cannot inspect raw grid cells, introduce arbitrary absolute coordinates, access planner costs or search parameters, or output a trajectory directly. This separation gives the LLM compositional flexibility while keeping map-dependent entity selection and geometry within a deterministic interface.

\subsection{Route Semantic Intermediate Representation}
\label{sec:route_specification}

Existing language-to-planning methods often use LTL~\cite{lang2ltl,lang2tlt2,ltlcodegen}, but Map2Route requires heterogeneous grounded route semantics and explicit scopes. We therefore introduce the \emph{Route Semantic Intermediate Representation (RouteIR)}:
\begin{equation}
    RouteIR=(\mathcal{S},\mathcal{C})
    \quad
    \mathcal{S}=(S_1,\ldots,S_J)
    \quad
    \mathcal{C}=(C_1,\ldots,C_L)
\end{equation}
where $\mathcal{S}$ contains ordered grounded destinations, and $\mathcal{C}$ the grounded route requirements. Each constraint is
\begin{equation}
    C_i=(\phi_i,\sigma_i),
    \qquad
    \sigma_i=(Q_i^{\mathrm{tmp}},Q_i^{\mathrm{spa}}),
\end{equation}
where $Q_i^{\mathrm{tmp}}$ and $Q_i^{\mathrm{spa}}$ specify its route-stage and spatial scopes, respectively, and the grounded semantic content is
$\phi_i \in \{
\operatorname{Avoid}(R),\allowbreak\;
\operatorname{Near}(R),\allowbreak\;
\operatorname{Far}(R),\allowbreak\;
\operatorname{Relative}(R_a,R_b),\allowbreak\;
\operatorname{PathShape}(W_1,\ldots,W_m)
\}$. $R$ denotes a grounded entity or region, and $(W_1,\ldots,W_m)$ an ordered waypoint sequence. Clearance is handled as a global planner regularizer.

\subsection{Grounding Verification and Bounded Repair}
\label{sec:verification}

Execution alone does not guarantee that the resulting specification is structurally valid or consistent with the semantic map. Grounding2Route therefore verifies reference validity and cardinality, constraint signatures, segment and scope consistency, and map-dependent relations by recomputing the corresponding queries through the semantic-map APIs. These checks detect execution, structural, and map-consistency errors, but do not independently certify semantic equivalence between the generated program and the original instruction.

When a verification error is detected, the verifier returns structured feedback $E^{(r)}$ for bounded LLM repair:
\begin{equation}
    P^{(r+1)}
    =
    f_{\mathrm{LLM}}
    \left(
        I,P^{(r)},E^{(r)}
    \right).
\end{equation}
This allows LLM to reconsider the complete grounding program against the instruction $I$ while correcting the detected error. The repaired program is re-executed and re-verified until verification succeeds or the repair budget is exhausted. 

\subsection{Scope-Aware Sequential Route Planning}
\label{sec:scope_planning}

Given the verified RouteIR, Grounding2Route plans each route segment sequentially using deterministic cost-aware A*. Going from $S_{k-1}$ to $S_{k}$, Path-Shape waypoints, and the grounded destination form an ordered target sequence $\{M_{k,1},\ldots,M_{k,m_k}\}$, with
\begin{equation}
    \tau_{k,j}
    =
    \operatorname{A}^{\ast}
    \left(M_{k,j-1},M_{k,j}\right),
\end{equation}
where consecutive trajectories are concatenated to form the complete route.

Hard avoidance requirements define segment-specific traversability masks, while Near, Far, and Relative preferences are converted into spatial costs. A constraint $C_i$ is active at segment $k$ and location $p$ only within its grounded route-stage and spatial scopes. The transition cost is
\begin{equation}
\begin{aligned}
c_k(p_t,p_{t+1})
&=
c_{\mathrm{geom}}(p_t,p_{t+1})
+
\sum_{i=1}^{L}
\lambda_{\phi_{i}} a_i(k,p_{t+1})J_{\phi_{i}}(p_{t+1})\\
&\quad+
\lambda_{\mathrm{clear}}J_{\mathrm{clear}}(p_{t+1}),
\end{aligned}
\end{equation}

\begin{equation}
\begin{aligned}
a_i(k,p)
&=
\mathbf{1}\!\left[
Q_i^{\mathrm{tmp}}=\varnothing
\vee k\in Q_i^{\mathrm{tmp}}
\right] \\
&\quad \cdot
\mathbf{1}\!\left[
Q_i^{\mathrm{spa}}=\varnothing
\vee p\in Q_i^{\mathrm{spa}}
\right].
\end{aligned}
\end{equation}
where $a_i$ is the indicator, $c_{\mathrm{geom}}$ is the standard 8-connected step cost, $\lambda$ is weight for each soft preferences.. We use
$J_{\mathrm{near}}(p)=\operatorname{clip}\left(d_{R_i}(p)/r_{\mathrm{near}},0,1\right)$,
$J_{\mathrm{far}}(p)=\exp\left(-d_{R_i}(p)/r_{\mathrm{far}}\right)$,
$J_{\mathrm{relative}}(p)=d_{\mathrm{close}}(p)/(d_{\mathrm{close}}(p)+d_{\mathrm{far}}(p)+\varepsilon)$,
and
$J_{\mathrm{clear}}(p)=\left([r_{\mathrm{clear}}-d_{\mathrm{obs}}(p)]+/r{\mathrm{clear}}\right)^2$.
We set $r_{\mathrm{near}}=r_{\mathrm{far}}=1.5,\mathrm{m}$ and $r_{\mathrm{clear}}=0.75,\mathrm{m}$. Clearance is applied globally.

Here, $d_{R_i}$ is the distance to reference region $R_i$, $d_{\mathrm{close}}$ and $d_{\mathrm{far}}$ to the closer/farther references, and $d_{\mathrm{obs}}$ to the nearest obstacle. All weights and distance scales are fixed across scenes. If any stage is infeasible, planning terminates without modifying RouteIR or triggering further LLM repair.

\section{Experiment Results}
We conduct experiments on Map2Route to answer four questions:
\textbf{(1)} How challenging is compositional language-grounded route planning for existing methods, and how does Grounding2Route compare with representative baselines (Section \ref{sec:ex_main})?
% \textbf{(2)} Which Grounding2Route design (controlled and task-specific adaptations) choices are critical to performance (Section~\ref{sec:ex_ablation})?
\textbf{(2)} Which task-specific components of Grounding2Route are most critical to performance, and how does each component contribute to Map2Route (Section~\ref{sec:ex_ablation})?
\textbf{(3)} What factors contribute to the difficulty of Map2Route, and how does performance vary with benchmark complexity (Section~\ref{sec:bench_analysis})?
\textbf{(4)} Can Grounding2Route execute compositional language-specified routes in real-world environments (Section \ref{sec:real_world})?

\subsection{Experiment Setup}

\textbf{Evaluation protocol:}
We evaluate all methods on the full Map2Route benchmark and report results separately on \textit{Easy} and \textit{Hard} episodes. We use Gemini-3.5-Flash for all LLM-based methods. For Grounding2Route, we set $(\lambda_{\mathrm{near}}, \lambda_{\mathrm{far}}, \lambda_{\mathrm{relative}}, \lambda_{\mathrm{clear}})=(8,96,64,36)$, tuned on the development set and fixed for all benchmark evaluations.

\textbf{Real-World Experiments:} We follow the SysNav~\cite{sysnav} pipeline to construct the semantic map. Experiments are conducted on a Mecanum-wheeled mobile robot~\cite{mecanum} equipped with a Livox Mid-360 LiDAR and a Ricoh Theta Z1 panoramic camera. A local collision-avoidance planner and a path follower are used to execute the globally planned route.
The system runs on a laptop equipped with an Intel Core i9-14900HX CPU and an RTX4090.

\textbf{Baseline adaptation:}
We compare Grounding2Route with seven representative methods: Lang2LTL~\cite{lang2ltl}, SayPlan~\cite{sayplan}, OSG-LLM~\cite{osgllm}, Lang2LTL-2~\cite{lang2tlt2}, LIMP~\cite{limp}, LTLCodeGen~\cite{ltlcodegen}, and ILN~\cite{iln}. Since no prior method natively supports the full Map2Route formulation, we adapt each method while preserving its original reasoning representation and planning pipeline. All methods receive the same benchmark-provided semantic map, traversability grid, instruction, and start pose, while retaining their native map abstractions and planners. 
% No baseline uses Grounding2Route's semantic-query APIs or ground-truth annotations, and no task-specific handling is added for unsupported soft preferences. 
This comparison reflects performance under the unified Map2Route setting rather than direct performance on the methods' original tasks.

\begin{table*}[h]
\centering
\caption{
Performance comparison on Map2Route across Easy and Hard difficulty levels.
% HCS evaluates hard-constraint satisfaction. SCS is decomposed into five soft-constraint categories: Near, Far, Relative, Path-Shape, and Clearance. Path efficiency is reported using raw path length (PL) and Success weighted by Path Length (SPL).
}
\label{tab:sempathbench_results}

\resizebox{0.9\textwidth}{!}{
\begin{tabular}{l|c|ccccc|cc}
\toprule

\multirow{2}{*}{\textbf{Method}}
& \multirow{2}{*}{\textbf{HCS} $\uparrow$}
& \multicolumn{5}{c|}{\textbf{Soft Constraint Metric}}
& \multicolumn{2}{c}{\textbf{Path Efficiency}} \\

\cmidrule(lr){3-7}
\cmidrule(lr){8-9}

&
&
\textbf{Near} $\downarrow$
& \textbf{Far} $\uparrow$
& \textbf{Relative} $\uparrow$
& \textbf{Path-Shape} $\uparrow$
& \textbf{Clearance} $\uparrow$
& \textbf{PL}
& \textbf{SPL} $\uparrow$ \\

\midrule

\multicolumn{9}{c}{\textbf{Easy}} \\
\midrule

\textbf{Shortest Feasible A*}
& 1.000
& 3.13 & 2.30 & 1.10 & 0.05 & 0.53 
& 28.0 & 1.00 \\

\textbf{Human Demonstration}
& 1.000
& 1.72 & 3.71 & 2.57 & 0.99 & 0.73
& 37.1 & 0.77 \\

\midrule

\textbf{Lang2LTL} \cite{lang2ltl} {\conf{CoRL23}}
& 0.300
& 4.41 & 0.76 & 0.49 & 0.01 & 0.51
& 30.1 & 0.28 \\

\textbf{SayPlan} \cite{sayplan} {\conf{CoRL23}}
& 0.102
& 4.74 & 0.35 & 0.13 & 0.00 & 0.52
& 20.2 & 0.08 \\

\textbf{OSG-LLM} \cite{osgllm} {\conf{ICRA24}}
& 0.368
& 4.66 & 0.93 & 0.57 & 0.02 & 0.46
& 21.6 & 0.36 \\

\textbf{Lang2LTL-2} \cite{lang2tlt2} {\conf{IROS24}}
& 0.248
& 4.45 & 0.63 & 0.42 & 0.02 & 0.51
& 31.3 & 0.22 \\

\textbf{LIMP} \cite{limp} {\conf{ICRA25}}
& 0.050
& 4.90 & 0.19 & 0.11 &  0.00 & 0.48
& 11.8 & 0.04 \\

\textbf{LTLCodeGen} \cite{ltlcodegen} {\conf{IROS25}}
& 0.274
& 6.03 & 0.65 & 0.38 & 0.02 & 0.49
& 26.2 & 0.26 \\

\textbf{ILN} \cite{iln} {\conf{IROS25}}
& 0.063
& 4.76 & 0.24 & 0.07 & 0.00 & 0.64
& 21.2 & 0.06 \\

\midrule

\textbf{Grounding2Route (Ours)}
& 0.667
& 3.21 & 2.76 & 1.27 & 0.16 & 0.74
& 31.2 & 0.53 \\

\textcolor{gainblue}{\scriptsize\textit{$\Delta$ vs. best baseline}}
& \gain{$+0.299$}
& \gain{$-1.20$}
& \gain{$+1.83$}
& \gain{$+0.70$}
& \gain{$+0.14$}
& \gain{$+0.10$}
& \gain{--}
& \gain{$+0.17$} \\

% \textcolor{gainblue}{\scriptsize\textit{vs. best baseline}}
% & \gain{$1.81\times$}
% & \gain{$1.37\times$}
% & \gain{$2.97\times$}
% & \gain{$2.23\times$}
% & \gain{$8.00\times$}
% & \gain{$1.16\times$}
% & \gain{--}
% & \gain{$1.47\times$} \\

\midrule
\midrule

\multicolumn{9}{c}{\textbf{Hard}} \\
\midrule

\textbf{Shortest Feasible A*}
& 1.000
& 2.99 & 2.41 & 1.24 & 0.05 & 0.53
& 66.2 & 1.00 \\

\textbf{Human Demonstration}
& 1.000
& 1.80 & 3.59 & 2.21 & 0.99 & 0.73
& 88.2 & 0.76 \\

\midrule

\textbf{Lang2LTL} \cite{lang2ltl} {\conf{CoRL23}}
& 0.124
& 4.61 & 0.41 & 0.16 & 0.00 & 0.51
& 72.7 & 0.10 \\

\textbf{SayPlan} \cite{sayplan} {\conf{CoRL23}}
& 0.029
& 4.83 & 0.14 & 0.05 & 0.00 & 0.52
& 44.8 & 0.03 \\

\textbf{OSG-LLM} \cite{osgllm} {\conf{ICRA24}}
& 0.120
& 5.56 & 0.38 & 0.15 & 0.00 & 0.35
& 31.8 & 0.11 \\

\textbf{Lang2LTL-2} \cite{lang2tlt2} {\conf{IROS24}}
& 0.094
& 4.68 & 0.29 & 0.09 & 0.01 & 0.51
& 53.7 & 0.08\\

\textbf{LIMP} \cite{limp} {\conf{ICRA25}}
& 0.012
& 4.96 & 0.05 & 0.00 & 0.00 & 0.43
& 13.2 & 0.01 \\

\textbf{LTLCodeGen} \cite{ltlcodegen} {\conf{IROS25}}
& 0.101
& 6.04 & 0.34 & 0.16 & 0.01 & 0.45
& 57.6 & 0.09 \\

\textbf{ILN} \cite{iln} {\conf{IROS25}}
& 0.011
& 4.91 & 0.07 & 0.02 & 0.00 & 0.65
& 19.0 & 0.01 \\

\midrule

\textbf{Grounding2Route (Ours)}
& 0.375
& 3.33 & 2.24 & 0.89 & 0.08 & 0.69
& 54.7 & 0.29 \\

\textcolor{gainblue}{\scriptsize\textit{$\Delta$ vs. best baseline}}
& \gain{$+0.251$}
& \gain{$-1.28$}
& \gain{$+1.83$}
& \gain{$+0.73$}
& \gain{$+0.07$}
& \gain{$+0.04$}
& \gain{--}
& \gain{$+0.18$} \\

%         \textcolor{gainblue}{\scriptsize\textit{vs. best baseline}}
% & \gain{$3.02\times$}
% & \gain{$1.38\times$}
% & \gain{$5.46\times$}
% & \gain{$5.56\times$}
% & \gain{$8.00\times$}
% & \gain{$1.06\times$}
% & \gain{--}
% & \gain{$2.64\times$} \\

\bottomrule
\end{tabular}
}
\vspace{-5mm}
\end{table*}

\subsection{Main Benchmark Performance} \label{sec:ex_main}

Table~\ref{tab:sempathbench_results} compares seven baselines and Grounding2Route with shortest-feasible A* and human demonstrations. The best baseline achieves only 0.368/0.124 HCS on Easy/Hard, revealing substantial difficulty in resolving compositional references. Soft constraints are similarly challenging, with best Path-Shape scores of only 0.02/0.01 versus 1.00 for humans, demonstrating the limited ability of current methods to satisfy soft constraints.

Grounding2Route substantially improves performance across all three evaluation dimensions, achieving 0.667/0.375 HCS and 0.53/0.29 SPL on Easy/Hard. It also outperforms all seven baselines across the five soft-preference metrics, showing stronger soft-constraint satisfaction. Despite these gains, a large gap to human performance remains: on Hard, Grounding2Route achieves 0.375 HCS and 0.08 Path-Shape, compared with 1.00 and 0.99 for human demonstrations.

\begin{table*}[t]
\centering
\caption{
Ablation study of Grounding2Route on Map2Route.
Each row modifies one component of the full Grounding2Route configuration
while keeping all remaining components fixed.
}
\label{tab:Grounding2Route_ablation}

\resizebox{0.95\textwidth}{!}{
\begin{tabular}{cccc|c|ccccc|cc}
\toprule

\multicolumn{4}{c|}{\textbf{Grounding2Route Configuration}}
& \multirow{2}{*}{\textbf{HCS} $\uparrow$}
& \multicolumn{5}{c|}{\textbf{Soft Preference Metrics}}
& \multicolumn{2}{c}{\textbf{Path Efficiency}} \\

\cmidrule(lr){1-4}
\cmidrule(lr){6-10}
\cmidrule(lr){11-12}

\textbf{Grounding}
& \textbf{IR}
& \textbf{Repair}
& \textbf{Soft Cons.}
&
& \textbf{Near} $\downarrow$
& \textbf{Far} $\uparrow$
& \textbf{Relative} $\uparrow$
& \textbf{Path-Shape} $\uparrow$
& \textbf{Clearance} $\uparrow$
& \textbf{PL} $\downarrow$
& \textbf{SPL} $\uparrow$ \\

\midrule

% ==================== Full Model ====================
CaG
& RouteIR
& 3
& Full
& \textbf{0.579}
& \textbf{3.27}
& \textbf{2.52}
& \textbf{1.11}
& \textbf{0.11}
& \textbf{0.72}
& \textbf{38.2}
& \textbf{0.46} \\

\midrule

% ==================== Grounding ====================
Schema
& RouteIR
& 3
& Full
& 0.338
& 4.25
& 1.05
& 0.38
& 0.06
& 0.60
& 19.6
& 0.27 \\

Direct ID
& RouteIR
& 3
& Full
& 0.474
& 3.91
& 1.45
& 0.84
& 0.01
& 0.69
& 28.2
& 0.39 \\

ToolCall
& RouteIR
& 3
& Full
& 0.539
& 3.84
& 1.67
& 0.84
& 0.02
& 0.66
& 27.2
& 0.44 \\

\midrule

% ==================== Intermediate Representation ====================
CaG
& LTL
& 3
& Full
& 0.439
& 3.92
& 1.10
& 0.70
& 0.03
& 0.65
& 24.3
& 0.36 \\

\midrule

% ==================== Repair ====================
CaG
& RouteIR
& 0
& Full
& 0.395
& 4.21
& 1.15
& 0.56
& 0.05
& 0.61
& 21.3
& 0.32 \\

CaG
& RouteIR
& 1
& Full
& 0.506
& 4.82
& 1.58
& 0.74
& 0.08
& 0.66
& 30.7
& 0.41 \\

CaG
& RouteIR
& 2
& Full
& 0.566
& 3.69
& 1.79
& 0.84
& 0.08
& 0.71
& 36.2
& 0.45 \\

\midrule

% ==================== Constraint / Scope ====================
CaG
& RouteIR
& 3
& Global-T
& 0.577
& 3.70
& 1.85
& 0.85
& 0.06
& 0.71
& 42.0
& 0.45 \\

CaG
& RouteIR
& 3
& Global-S
& 0.580
& 3.67
& 1.91
& 0.87
& 0.10
& 0.71
& 37.9
& 0.47 \\

CaG
& RouteIR
& 3
& Global-TS
& 0.577
& 3.70
& 1.87
& 0.84
& 0.07
& 0.70
& 41.1
& 0.46 \\

CaG
& RouteIR
& 3
& Hard Only
& 0.581
& 3.79
& 1.36
& 0.58
& 0.04
& 0.48
& 31.2
& 0.56 \\

\bottomrule
\end{tabular}
}
\end{table*}

\begin{figure*}[h]
\centering
\includegraphics[width=0.95\linewidth]{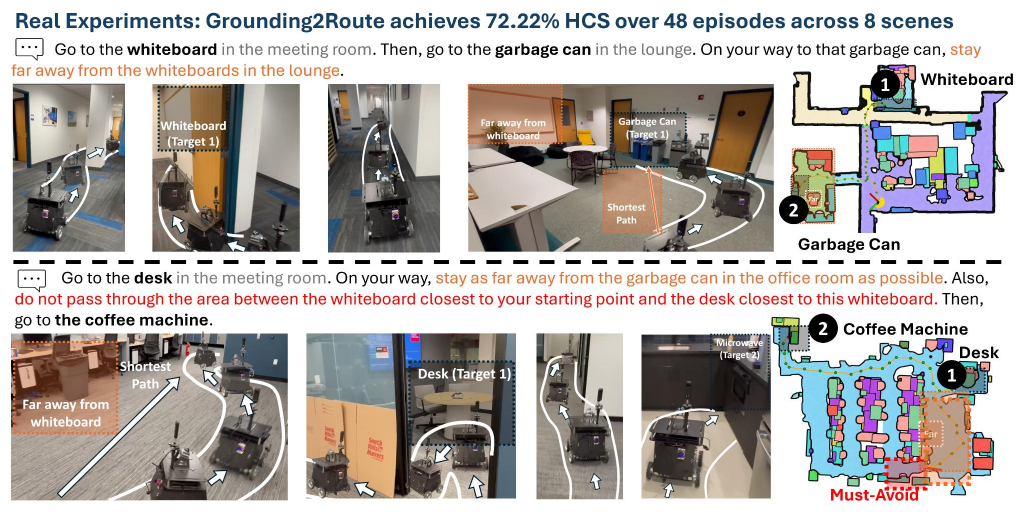}
\caption{
Qualitative real-world evaluation of Grounding2Route. Two representative episodes demonstrate multi-stage target grounding and execution under language-specified route constraints, including scoped far preferences and must-avoid requirements. 
The executed trajectories and semantic maps demonstrate successful execution of compositional route instructions indoors.
}
\vspace{-5mm}
\label{fig:realworld}
\end{figure*}

\subsection{Method Ablation Study}  \label{sec:ex_ablation}

Unlike the minimally adapted methods in the main benchmark comparison, the following ablations use controlled, task-specific variants to isolate individual design choices in Grounding2Route. Table~\ref{tab:Grounding2Route_ablation} evaluates each key compnent design choices while keeping the remaining pipeline fixed.

\subsubsection{Grounding strategy}
We compare CaG with two direct-prediction variants and an iterative tool-calling variant. \emph{Schema} predicts a schema-constrained JSON intent graph that is
deterministically compiled for grounding, while \emph{Direct ID} selects referenced entity IDs from a simplified map representation. \emph{ToolCall} uses the same semantic APIs and downstream pipeline as CaG, but invokes the APIs iteratively through LLM function calls instead of generating an executable grounding program. Schema, Direct ID, and ToolCall achieve HCS/SPL of 0.338/0.27, 0.474/0.39, and 0.539/0.44, respectively, compared with 0.579/0.46 for CaG. This controlled comparison shows that executable program composition is more effective than iterative tool calling for compositional route grounding.

\subsubsection{Intermediate representation}
Replacing RouteIR with LTL reduces HCS/SPL from 0.579/0.46 to 0.439/0.36, showing that RouteIR is better suited to representing the heterogeneous route semantics and scopes in Map2Route.

\subsubsection{Verification and repair} We vary the maximum number of verification-guided repair rounds. Increasing the budget progressively improves HCS from 0.395 at $R=0$ to 0.506, 0.566, and 0.579 at $R=1,2,3$, respectively. We also observe improvements across all five soft-constraint metrics, with Relative increasing from 0.56 to 1.11. These gains indicate that repair corrects not only hard requirements but also the grounding of soft preferences and their scopes.

\subsubsection{Soft-constraint modeling} We ablate both explicit scope modeling defined in RouteIR and preference-aware planning. We first globalize the temporal scope (Global-T), spatial scope (Global-S), or both (Global-TS), such that constraints remain active throughout the full route, the full map, or both. These variants leave HCS nearly unchanged but reduce Far from 2.52 to 1.85--1.91 and Relative from 1.11 to 0.84--0.87. We further remove all soft-preference costs while retaining the grounded destinations and hard constraints. This produces higher path efficiency but reduces all soft metrics. These results show explicitly modeling both soft preferences and their scopes is essential for language-faithful route planning.

\subsubsection{Planner over Oracle Grounding}

Since the soft metrics are coupled with HCS to avoid metric hacking, we provide ground-truth RouteIR to isolate our scope-aware planner.

\begin{wraptable}{r}{0.47\linewidth}
\vspace{-2mm}
\centering
\caption{Planner ablation over oracle grounding.}
\vspace{-0.8em}
\label{tab:oracle_decomposition}
\scriptsize
\setlength{\tabcolsep}{2.5pt}
\begin{tabular}{lccc}
\toprule
Planner & Near $\downarrow$ & Far $\uparrow$ & Rel. $\uparrow$ \\
\midrule
Shortest A* & 3.07 & 2.35 & 1.16 \\
Human Demo  & 1.76 & 3.65 & 2.42 \\
Ours        & 2.09 & 3.33 & 2.06 \\
\bottomrule
\end{tabular}
\vspace{-0.8em}
\end{wraptable}

From Table~\ref{tab:oracle_decomposition}, all targets are reached, and our planner attains 2.09 / 3.33 / 2.06 on Near / Far / Relative, consistently improving over shortest-path A* and recovering 74\% of the A*-to-human gap on average. These results suggest that grounding remains the dominant source of error in the full pipeline, while a smaller but non-negligible gap remains in soft-constraint-aware planning.

\vspace{-1mm}

\subsection{Benchmark Difficulty Analysis}  \label{sec:bench_analysis}

\begin{figure}[h]
\vspace{-5mm}
\centering
\includegraphics[width=\linewidth]{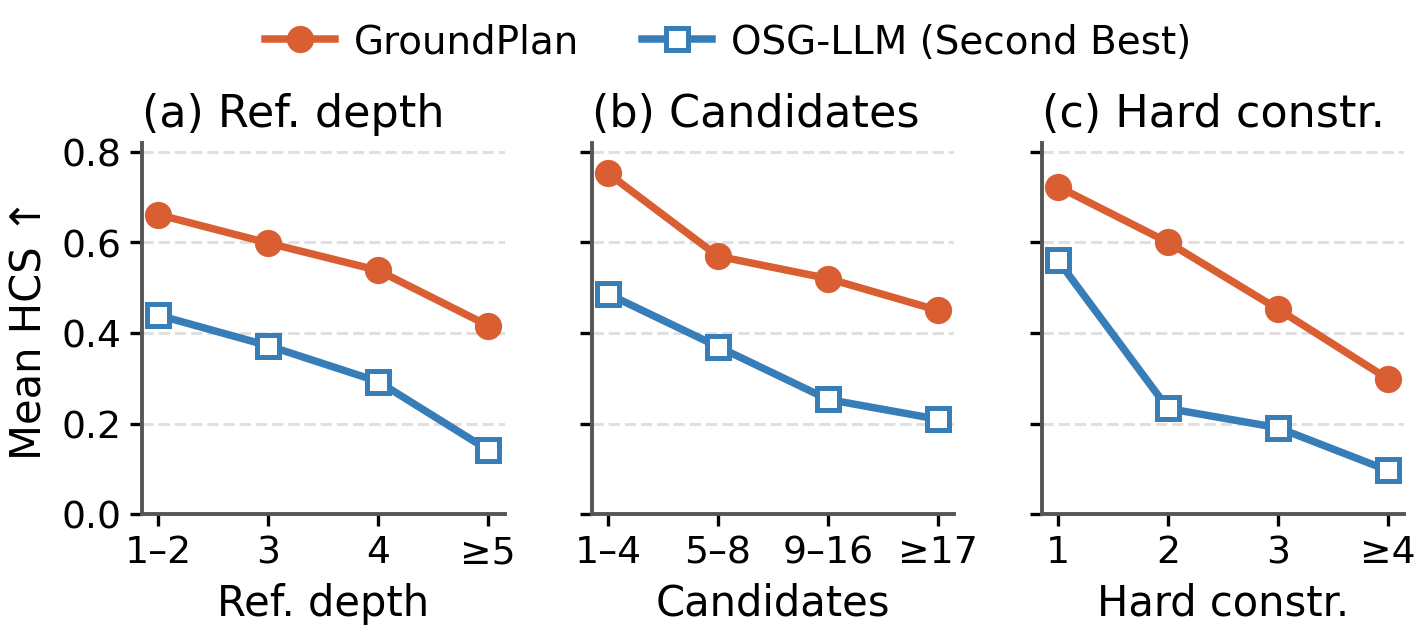}
\caption{
Difficulty analysis of Map2Route. Mean HCS of Grounding2Route and OSG-LLM across increasing reference depth, candidate instances and hard constraint number.
}
% \vspace{-6mm}
\label{fig:diff_analysis}
\end{figure}

Fig.~\ref{fig:diff_analysis} analyzes three complementary sources of benchmark difficulty: compositional reference depth, instance ambiguity, and hard-constraint complexity. HCS decreases for both Grounding2Route and OSG-LLM as each factor increases, with larger degradation under deeper references and more hard constraints. Grounding2Route remains consistently stronger across all complexity levels, but also declines in the hardest settings, highlighting the increasing difficulty of compositional grounding and route reasoning.

\subsection{Real-World Experiments} \label{sec:real_world}
We further evaluate Grounding2Route in the real world with 48 episodes across 8 indoor scenes. Grounding2Route achieves 72.22\% HCS. 
Among the HCS loss, 73.75\% result from semantic-map construction errors, 22.5\% from Grounding2Route failures in grounding or route planning, and 3.75\% from SLAM issues.
Qualitative results in Fig. \ref{fig:realworld} and \href{https://anonymous.4open.science/w/Map2Route-F05F/}{Project Page} show Grounding2Route's ability to achieve soft preferences during real-world navigation.

\section{Conclusion}

We introduced Map2Route, a benchmark for compositional language-grounded route planning over semantic maps, covering nested references, ordered hard requirements, and scoped soft preferences. 
Evaluation with seven adapted baselines and human demonstrations shows that current methods remain substantially limited, with performance degrading as grounding and route complexity increase. 
We further present Grounding2Route as a strong benchmark method, which consistently improves hard-requirement satisfaction, soft-preference adherence, and path efficiency, with additional validation in real-world environments.   
Nevertheless, the large remaining gap to human performance indicates that Map2Route is far from saturated and provides substantial headroom for future research.

\textbf{Limitations:}
1) Map2Route intentionally isolates grounding and planning from online perception through a pre-constructed semantic map. However, real-world deployment introduces additional scene-construction errors, which our robot experiments identify as a major source of failure.
2) The benchmark currently covers 40 scenes and is annotated by two experts, limiting scene diversity relative to large-scale cross-domain environments and potentially constraining the diversity of linguistic styles and annotation strategies.
3) The current benchmark does not consider dynamic or multi-floor settings, which we leave for future extension.

% \textbf{Limitations:}
% 1) Map2Route isolates grounding and planning from online perception, while real-world deployment remains affected by scene-construction errors.
% 2) With 40 scenes and two annotators, scene and language diversity remain limited.
% 3) Dynamic and multi-floor settings are left for future work.

% \section*{ACKNOWLEDGMENT}

% This is acknowledgement

\bibliographystyle{IEEEtran}
\bibliography{reference}

\end{document}